\documentclass[letterpaper, 10 pt, conference]{ieeeconf}

\IEEEoverridecommandlockouts
\usepackage{amsmath}
\usepackage{amssymb}
\usepackage{graphicx}
\usepackage{booktabs}
\usepackage{multirow}
\usepackage[table]{xcolor}
\usepackage{algorithm}
\usepackage{algpseudocode}
\usepackage{float}
\usepackage{cite}
\usepackage{hyperref}

\hypersetup{
    colorlinks=true,
    urlcolor=blue,
}

\providecommand{\citep}[1]{\cite{#1}}
\makeatletter
\let\driftworld@makecaption\@makecaption
\long\def\@makecaption#1#2{%
    \ifx\@captype\@IEEEtablestring
        \@IEEEfigurecaptionsepspace
        \setbox\@tempboxa\hbox{\footnotesize #1.\ #2}%
        \ifdim \wd\@tempboxa >\hsize
            \setbox\@tempboxa\hbox{\footnotesize #1.\ }%
            \parbox[t]{\hsize}{\footnotesize\noindent
                \parfillskip=0pt plus .4\hsize\relax
                \unhbox\@tempboxa#2}%
        \else
            \hbox to\hsize{\hfil\footnotesize\box\@tempboxa\hfil}%
        \fi
    \else
        \driftworld@makecaption{#1}{#2}%
    \fi
}
\makeatother

\newif\ifanon
\title{\LARGE \bf
DriftWorld: Fast World Modeling through Drifting
}

\ifanon
\author{Anonymous Authors}
\else
\author{%
Susie Lu$^{1}$,
Haonan Chen$^{2}$,
Weirui Ye$^{1}$,
and Yilun Du$^{2}$%
\thanks{$^{1}$Massachusetts Institute of Technology. $^{2}$Harvard University.}%
}
\fi

\begin{document}

\maketitle
\thispagestyle{empty}
\pagestyle{empty}

\begin{figure*}[t]
    \centering
    \includegraphics[width=0.95\linewidth]{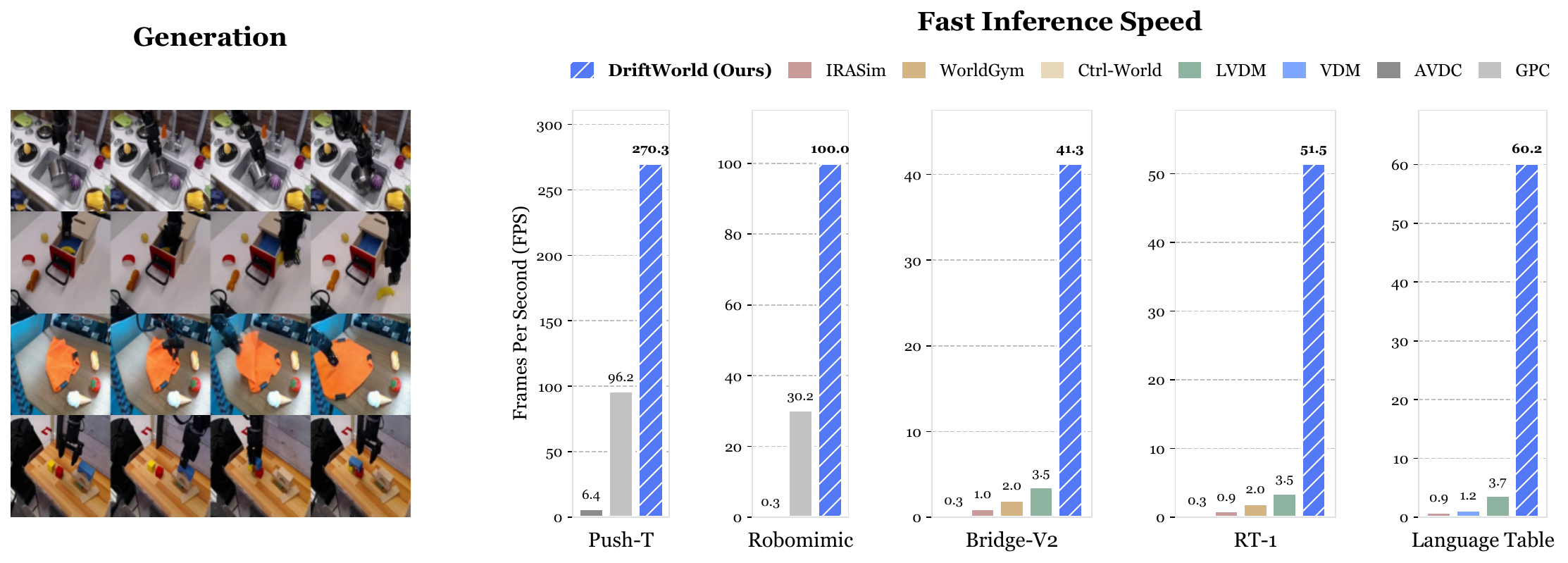}
    \vspace{-8pt}
    \caption{\textbf{Overview of DriftWorld.} DriftWorld is an action-conditioned world model based on drifting that generates future frames in a single forward pass. It achieves 40+ fps generation, which is significantly faster than existing models on all five environments. DriftWorld can be used for high-quality generation, efficient planning, and offline simulation of policies.}
    \label{fig:teaser}
    \vspace{-15pt}
\end{figure*}

\begin{abstract}
Predictive world models enable robots to simulate the visual outcomes of their actions, but state-of-the-art diffusion-based models remain costly because generating each rollout requires multi-step iterative denoising. We introduce DriftWorld, an action-conditioned world model based on \textit{drifting generative models}. DriftWorld learns a conditional drift during training, enabling it to generate future observations for a given action sequence in a single forward pass during inference. Across Bridge-V2, RT-1, Language Table, Push-T, and Robomimic, DriftWorld runs at over 40 fps and is 12+ times faster than diffusion-based baselines, while matching or improving their visual generation quality. This makes DriftWorld an efficient world model for robot simulation and further enables downstream applications including inference-time action search and offline policy evaluation. Videos and code are available at \href{https://susie-lu.github.io/driftworld/}{this website}.
\end{abstract}

\vspace{-10pt}
\section{Introduction}
\vspace{-3pt}

Predictive world models have emerged as powerful tools for robot learning, letting robots predict the outcomes of their actions without executing in the real world. With recent advances in video generation, action-conditioned world models can now simulate fine-grained robot--object interactions \cite{yang2024unisim,zhu2025irasim}, render controllable multi-view videos \cite{guo2026ctrl}, and produce long-horizon rollouts~\cite{du2024video}. Such models lead to two key applications. First, robots can improve their policies by rolling out candidate action proposals in the world model and selecting the best one to execute in the real world. Second, the model serves as an offline simulator for policy evaluation.

However, the practical value of these world models is bottlenecked by inference speed. State-of-the-art models are predominantly diffusion-based \cite{guo2026ctrl, qi2026inference}, so they rely on iterative, multi-step denoising to generate future frames. This multi-step sampling is too slow for real-time planning. For instance, recent work on generative predictive control \cite{qi2026inference} reports that diffusion world model rollouts consume 90--95\% of runtime, resulting in 3 or more seconds per decision cycle. As a result, it is impractical to simulate the hundreds of rollouts needed to identify the best candidate actions.

To overcome this bottleneck, we introduce DriftWorld, a fast action-conditioned world model based on drifting generative models \cite{deng2026generative}. Unlike diffusion-based models, DriftWorld generates high-quality future frames in a single forward pass. DriftWorld achieves 1-step generation by learning a drifting field at training time that maps the prior noise distribution to the data distribution. Thus, inference does not require the iterative, multi-step sampling used in diffusion models.

The original drifting model was designed for class-conditional image generation. Adapting it to action-conditioned video generation requires rethinking three components. We introduce: (1) a conditional drifting field defined by the initial observation and action sequence with a single-positive training objective; (2) feature-space integration using DINOv2/v3 \cite{oquab2023dinov2,simeoni2025dinov3} to maintain visual sharpness in complex scenes; and (3) action-accentuation to strengthen the generated video's adherence to the specified actions.

We evaluate DriftWorld on standard vision-based robotic manipulation benchmarks, including the real-world Bridge-V2 \cite{walke2023bridgedata}, RT-1 \cite{rt12022arxiv}, and Language Table \cite{lynch2023interactive} datasets and the simulated Push-T \cite{florence2022implicit,chi2025diffusion} and Robomimic \cite{mandlekar2021matters} environments. DriftWorld is 12+ times faster than diffusion world-model baselines while matching or improving the visual quality of generated rollouts. Using DriftWorld to evaluate policy action proposals before execution also improves control performance, increasing Push-T IoU from 0.635 to 0.772. Beyond online control, DriftWorld also serves as an offline simulator for ranking real-world policies, reaching Pearson correlation coefficients of 0.9755, 0.9958, and 0.9618 with ground-truth performance on Push-T, Robomimic Lift, and Robomimic Can. Our main contributions are:
\begin{itemize}
    \item \textbf{DriftWorld:} We introduce a fast action-conditioned world model that generates each rollout in a single forward pass.
    \item \textbf{Specializing Drifting for Robot Simulation:} We present the first adaptation of drifting generative models to conditional video generation, where we introduce a single-positive drifting objective, feature space integration, and an action-accentuated drifting field.
    \item \textbf{Experimental Validation:} We show that DriftWorld matches or exceeds baselines in rollout quality across five benchmarks while running significantly faster, which makes both inference-time action search and offline policy evaluation more efficient.
\end{itemize}
\section{Related Work}

\textbf{Action-Conditioned World Models for Robotics.} World models that predict future observations conditioned on actions have a long history in model-based reinforcement learning \cite{ha2018world,hafner2020dreamer} and in robotics \cite{lozano1983robot, hou2026world, chen2023predicting, hong2026scoop}. Early approaches operated in compact latent spaces, but recent work has shifted toward high-fidelity visual prediction by adapting video generation models. UniSim \cite{yang2024unisim}, GAIA-1 \cite{hu2023gaia1}, Genie \cite{bruce2024genie}, and Cosmos \cite{nvidia2025cosmos} treat the world as a video that unrolls in response to actions or text, and a growing line of work repurposes pretrained video diffusion models as interactive simulators \cite{he2025dws,rigter2025avid,huang2025vid2world}. Further, robotics-specific models, including IRASim \cite{zhu2025irasim}, Ctrl-World \cite{guo2026ctrl}, Veo-Robotics \cite{team2025evaluating}, and UWM \cite{zhu2025uwm}, produce fine-grained and multi-view manipulation rollouts, and they are increasingly used as environments for policy training and evaluation \cite{quevedo2025worldgym,team2025evaluating,wang2026interactiveworldsim}. However, these models predominantly rely on multi-step diffusion, requiring tens of forward passes per frame, which makes rollouts expensive. DriftWorld achieves comparable visual quality but replaces iterative sampling with an efficient, single-step drifting generator specialized for action-conditioned prediction.

\textbf{Fast and Few-Step Generative Models.} A growing body of work aims to reduce the sampling cost of diffusion models. Progressive distillation \cite{salimans2022progressive} and consistency models \cite{song2023consistency,song2024improved} train students to take large or single steps along the diffusion ODE; rectified flow \cite{liu2023flow} straightens probability paths to enable few-step sampling; and adversarial objectives such as ADD \cite{sauer2024adversarial} push one-step generation through distillation from a teacher. Related ideas have also been applied to the video setting \cite{gao2025magicdrivedit,lin2024animatediff}. Most of these methods assume a pretrained multi-step diffusion teacher. In contrast, drifting generative models \cite{deng2026generative} train a one-step generator from scratch by moving its pushforward distribution toward the data distribution using a kernelized attraction-repulsion field. Our work is the first, to our knowledge, to adapt drifting to a conditional video prediction problem. We show that the adaptations of a single-positive drifting objective, action accentuation, and feature-space drifting are essential to make drifting work well in this case.

\textbf{World Models for Planning and Policy Evaluation.} 
With a world model that simulates future dynamics, we can roll out candidate action sequences in imagination, enabling both online policy improvement and offline policy ranking. Sampling-based planners in latent world models \cite{hafner2020dreamer, hansen2022temporal} and generative predictive control \cite{qi2026inference} score many rollouts and execute the best one. However, as noted in \cite{qi2026inference}, diffusion rollouts account for most of the decision-time compute, which limits the number of proposals that can be evaluated. Alternatively, we can use world models to directly imagine future plans~\citep{du2023learning, kim2026cosmos, ye2026world}, but such approaches are also limited by the slow iterative nature of diffusion generation. For offline evaluation, Veo-Robotics \cite{team2025evaluating}, WorldGym \cite{quevedo2025worldgym}, and UniSim \cite{yang2024unisim} use a learned simulator to rank real-world policies without costly hardware rollouts. DriftWorld targets both settings: its one-step generation enables more efficient online action search, and its high-fidelity predictions make it a reliable offline simulator.

\section{Drifting World Model}

\begin{figure*}[t]
    \centering
    \includegraphics[width=\linewidth]{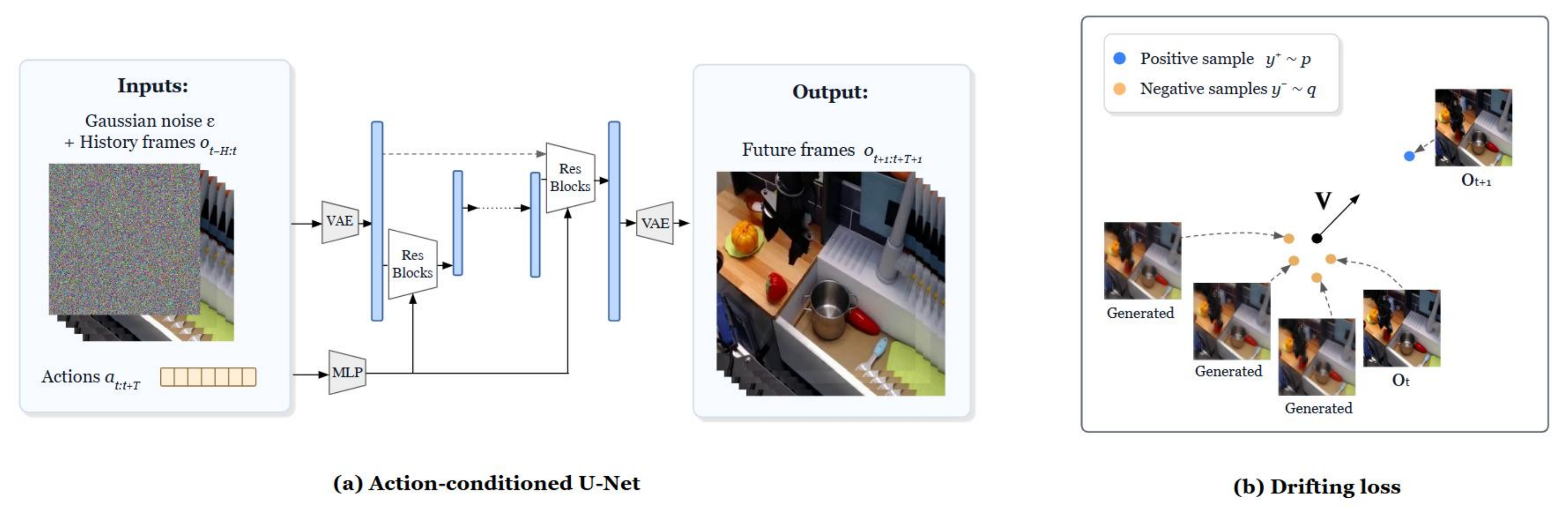}
    \vspace{-15pt}
    \caption{\textbf{Drifting Architecture and Training Overview.} (a) We use a U-Net architecture conditioned on history observations and robot actions. (b) DriftWorld is trained using a drifting loss, which is a contrastive loss that encourages the generated future frames (negative samples) to drift toward the ground-truth future frames (the positive sample).
    }
    \label{fig:driftworld}
    \vspace{-13pt}
\end{figure*}

In this section, we present DriftWorld, an action-conditioned world model that generates high-quality predictions of future frames in a single forward pass. We first formulate the action-conditioned world modeling problem and the drifting setup, and then introduce two key adaptations for conditional video generation: a single-positive drifting objective and action-conditioning and feature-space designs.

\subsection{Problem Formulation}
\vspace{-3pt}

We address the problem of action-conditioned video generation. Let $o_t$ denote a visual observation (such as a video frame) at time $t$, and let $a_t$ denote the action that the robot takes at time $t$. Given a history of past observations $o_{t-F:t} = (o_{t-F}, \dots, o_t)$ and a sequence of proposed actions $a_{t:t+T-1} = (a_t, \dots, a_{t+T-1})$, our objective is to accurately predict the resulting future observations $o_{t+1:t+T}$:
$$ o_{t+1:t+T} \sim \mathcal{W}(\cdot \mid o_{t-F:t}, a_{t:t+T-1}), $$
where $\mathcal{W}$ is the action-conditioned world model.

\subsection{One-Step Video Generation via Drifting}
\vspace{-3pt}

To apply the drifting method \cite{deng2026generative} to video generation, we formulate the process as follows. Let $f_\theta$ denote the drifting world model, which maps a noise prior $\epsilon$ and conditioning $c$ to the generated video frames $x = f_\theta(\epsilon, c)$, where $c$ consists of the history frames $o_{t-F:t}$ and future actions $a_{t:t+T-1}$, and $x$ is the predicted $o_{t+1:t+T}$.

Let $p$ denote the true data distribution, and let $q$ denote the distribution of model outputs, i.e. $x \sim q$. While diffusion models iteratively refine $q$ toward to $p$ during \emph{inference time}, drifting models optimize $q$ to evolve towards $p$ during \emph{training time} so that it can perform inference in a single forward pass. This evolution is driven by a drifting field $V_{p,q}$. Specifically, let $q_i$ be the model output distribution at the $i$-th training step, and consider a generated sample $x_i \sim q_i$. At the $(i+1)$-st training step, the sample will become $x_{i+1} = x_i + V_{p, q_i}(x_i).$ The drift vector $V_{p, q_i}(x_i)$ moves $x_i$ to $x_{i+1}$ by attracting $x_i$ toward data samples in $p$ (known as positive samples) and repelling it away from generated samples in $q_i$ (negative samples). As the model output distribution becomes closer to the true data distribution, the drift becomes smaller and eventually close to zero.

The model $f_\theta$ is trained by minimizing the distance between the model output $x_i$ and its drifted target $x_{i+1}$:
\begin{equation}\label{eq:loss}
\text{Loss} = \mathbb{E}_{\epsilon} \left[ \| x_i - \text{stopgrad}(x_{i+1}) \|^2 \right]
\end{equation}
where $x_i = f_{\theta_i}(\epsilon, c)$.

To handle action-conditioned video generation, we introduce two key components that differ from the original image-generation drifting \cite{deng2026generative}: (i) a single-positive drifting field (Section \ref{sec:drift-1pos}) and (ii) an action-conditioning mechanism and feature space for video generation (Section \ref{sec:video}).

\subsection{Design of Single-Positive Drifting}\label{sec:drift-1pos}
\vspace{-3pt}

\textbf{Positive and Negative Samples.} While a class-conditional image generation model can use multiple positive samples drawn from each class, DriftWorld can only use a single positive sample, as the training dataset only provides one sequence of future frames given the history frames and proposed actions. As a result, DriftWorld operates on (i) a single positive sample, which is the ground-truth future frames $o_{t+1:t+T}$, and (ii) $N$ negative samples, which are future frames $\hat{o}_{t+1:t+T}$ generated by the current model.

\textbf{Computing the Drifting Field.} Next, we derive the expression for computing the drifting field from the single positive sample and $N$ negative samples. Consider the general equation \cite{deng2026generative} for the drifting field on a particular model output $x$:
\[ V_{p,q}(x) = \mathbb E_{y^+\sim p, y^-\sim q}\big[\Tilde{k}(x,y^+)\Tilde{k}(x,y^-)(y^+ - y^-)\big], \]
where $y^+$ represents a positive sample from the true data distribution $p$, and $y^-$ represents the $N$ negative samples from the current model's output distribution $q$. The function $\Tilde{k}(\cdot,\cdot)$ is a normalized kernel that quantifies the similarity between two samples. In practice, during each training step, the model generates $N$ samples. We evaluate the drifting field at each generated sample in turn by treating it as $x$ and using the remaining generated samples as negatives $y^-$. Thus, we end up with $N$ values of $V_{p,q}(x)$, which we plug into \eqref{eq:loss} and take the average to obtain the final drifting loss.

Since there is only a single positive sample, the expression for $V_{p,q}(x)$ becomes
\begin{equation*}
\resizebox{\columnwidth}{!}{$
\begin{aligned}
V_{p,q}(x) &= \Tilde{k}(x,y^+) \mathbb E_{y^-\sim q}\big[\Tilde{k}(x,y^-)(y^+ - y^-)\big] \\
&= \Tilde{k}(x,y^+) \Big(\mathbb E_{y^-\sim q}\big[\Tilde{k}(x,y^-)y^+\big] - \mathbb E_{y^-\sim q}\big[\Tilde{k}(x,y^-)y^-\big] \Big) \\
&= \Tilde{k}(x,y^+) \Big(\mathbb E_{y^-\sim q}\big[\Tilde{k}(x,y^-)\big]y^+ - \mathbb E_{y^-\sim q}\big[\Tilde{k}(x,y^-)y^-\big] \Big)
\end{aligned}
$}
\end{equation*}
Let $Z = \mathbb E_{y^-\sim q}[\Tilde{k}(x,y^-)]$ and $\hat{y}^- = \frac{1}{Z} \mathbb E_{y^-\sim q}[\Tilde{k}(x,y^-) y^-]$ denote the kernel-weighted center of the negative samples around $x$. Then
\begin{equation}\label{eq:1pos-new}
V_{p,q}(x) = \Tilde{k}(x,y^+) \cdot Z \cdot (y^+ - \hat{y}^-) 
\end{equation}
where $\Tilde{k}(x,y^+) \cdot Z$ is a scalar.

Equation \eqref{eq:1pos-new} states that the drift is a scalar multiple of $y^+ - \hat{y}^-$, which is a vector pointing from the kernel-weighted center of the model's output distribution toward the ground-truth target. The purpose of kernel weighting is as follows. Consider the drifting update on a sample $x$: we want $x$ to be mostly attracted or repelled by samples that are sufficiently close to $x$, rather than by samples that are distant or in different modes of the output distribution. Kernel weighting in the terms $\Tilde{k}(x,y^+)$ and $\hat{y}^-$ of \eqref{eq:1pos-new} helps achieve this by putting a higher weight on samples closer to $x$.

It is worth noting that DriftWorld's training objective is different from Mean Squared Error (MSE) loss. MSE loss can be viewed as a special case of \eqref{eq:loss}, where $V_{p,q}(x_i) = y^+ - x_i$ so that $x_{i+1} = y^+$ and the loss equals $\|x_i - y^+\|^2$. In contrast, DriftWorld's training objective is formed by substituting the single-positive drifting field \eqref{eq:1pos-new} into the loss \eqref{eq:loss}, so the update for each $x_i$ depends on the other generated samples through $\hat{y}^-$. Thus, while MSE moves each prediction independently toward the ground truth, drifting defines each update using the model's generated distribution, approximated by the $N-1$ other generated samples.

\subsection{Design of Action Conditioning and Feature Space}\label{sec:video}
\vspace{-3pt}

Besides the single-positive training objective, we introduce action accentuation to prevent the model from ignoring the action and simply predicting little change from the current observation. We also use feature-space drifting so that the drift is computed using a kernel with a more meaningful similarity measure for complex real-world scenes.

\textbf{Accentuating Action Following.} To prevent the model from ignoring the action conditioning, we add the current ground-truth frame $o_t$ as an additional negative sample. Then, the drifting loss is computed with a total of $N+1$ negative samples, which are $o_t$ and the $N$ generated samples, and a single positive sample, which is the ground-truth $o_{t+1}$. Adding $o_t$ as a negative contributes a component of the drift \eqref{eq:1pos-new} in the direction of $o_{t+1} - o_t$: the drift pushes predictions away from the current observation $o_t$ and toward the ground-truth future observation $o_{t+1}$, which emphasizes the visual change caused by taking the action. 

Each generated negative has weight 1, while $o_t$ has weight $(\alpha-1)(N-1)$, where $\alpha\geq1$ is the accentuation scale. When computing the drift for a generated sample, the $N-1$ other generated samples act as negatives; thus, the factor $N-1$ puts the contribution of $o_t$ on the same scale as these generated negatives as a group. When $\alpha=1$, $o_t$ has zero weight, while increasing $\alpha$ strengthens its contribution to the drifting field. This approach can be viewed as a form of the accentuation mechanism in \cite{deng2026generative}. During training, we sample $\alpha$ from a log-uniform distribution, e.g. over $[1,4]$, and condition the U-Net on its value. This allows a single trained model to support different accentuation strengths at inference time, so we can sweep the values of $\alpha$ to select the value that performs best.

\textbf{Feature Space for Drifting.} For complex real-world scenes, we adapt the drifting loss in two ways: (i) compute it in feature space so that the kernel uses a more meaningful similarity measure, and (ii) put a larger weight on moving regions to prevent static backgrounds from dominating the training objective.

The kernel used to compute the drifting field relies on pairwise similarity between samples. For simpler environments, we compute the drifting loss directly in the model's native output space, such as pixel space. For complex real-world scenes, however, distances in the model's native output space may not provide a meaningful measure of similarity. Thus, we project the model outputs and ground truth into a feature space using an encoder $\phi$ and compute the drifting loss there. This encourages $\phi(f_\theta(\epsilon, c))$ to drift toward $\phi(y^+)$ and away from $\phi(y^-)$. Since the feature encoder is only used to compute the loss during training, it does not affect the model's inference speed.

In our implementation, $\phi$ is DINOv3 \cite{simeoni2025dinov3}, and each video frame is encoded independently. Each frame is divided into $16 \times 16$ patches, and each patch is encoded into a $D$-dimensional feature. We compute the drifting field independently for each patch, and sum the losses over patches.

If all patches are weighted equally, static background patches can dominate the aggregate drifting loss, while patches corresponding to smaller regions with motion contribute relatively little. To address this issue, we assign greater weight to the drifting loss on patches whose features change more between the current and target future frames. Specifically, the motion weight is the hyperbolic tangent of the L2 distance between a patch's features in the target future frame(s) and the current frame $o_t$.

\begin{algorithm}[t]
\caption{\textbf{Training Step}}
\label{alg:train}
\renewcommand{\algorithmicrequire}{\textbf{Input:}}
\begin{algorithmic}[1]
\Require $f_\theta$: U-Net model
\Require $y^+$: positive sample
\Require $o_{t-F:t}$: current and history observations
\Require $a_{t:t+T-1}$: proposed actions
\Require $N$: number of generated samples for drifting
\State $\epsilon \gets \{\epsilon_i \sim \mathcal{N}(0,I)\}_{i=1}^N$ \Comment{noise}
\State $x \gets f_\theta(\epsilon, o_{t-F:t}, a_{t:t+T-1})$ \Comment{generated samples}
\State $y^- \gets \text{cat}([x, o_t])$ \Comment{negative samples}
\State $V \gets \text{compute\_V}(x, y^+, y^-)$
\State $\text{loss} \gets \|x - \text{stopgrad}(x + V)\|^2$
\end{algorithmic}
\end{algorithm}

\vspace{-2pt}
\subsection{Action-Conditioned U-Net Architecture}
\vspace{-3pt}

Algorithm \ref{alg:train} provides the pseudocode for a single training step. DriftWorld uses an action-conditioned U-Net (Fig. \ref{fig:driftworld}a) that takes the current and history frames $o_{t-F:t}$ and the actions $a_{t:t+T-1}$ as input and predicts future frames $o_{t+1:t+T}$. To generate multiple future frames at once, we use factorized spatial-temporal convolutions \cite{ko2024learning}: a spatial convolution is applied independently and identically to each of the $T$ time steps, and then a temporal convolution is applied independently and identically at each spatial location.

Action conditioning is incorporated frame-wise through FiLM \cite{perez2018film}, with each action $a_{t+i}$ corresponding to the future frame $o_{t+i+1}$ that results from taking the action. The model also conditions on history frames by concatenating them channel-wise with the initial Gaussian noise.

\vspace{-3pt}
\subsection{Inference and Policy Simulation}
\vspace{-3pt}

At inference time, DriftWorld generates future video frames using a single forward pass $f_\theta(\epsilon, c)$, where $\epsilon$ is Gaussian noise and $c = (o_{t-F:t}, a_{t:t+T-1})$.

This enables efficient downstream policy simulation in DriftWorld. Consider a vision-based policy $\pi$ that receives a history of visual observations and outputs a chunk of actions $a_{t:t+T-1}$ to execute. The world model simulates the outcomes of the actions, and the final predicted observation $o_{t+T}$ can be fed back into the policy $\pi$ to generate the next action chunk. Continuing this process autoregressively, we can simulate long-horizon policy rollouts in the world model.

\vspace{-10pt}
\section{Experiments}\label{sec:exp}
\vspace{-3pt}

In this section, we conduct experiments to evaluate DriftWorld, demonstrating that it matches or exceeds the performance of existing action-conditioned world models while generating rollouts significantly faster. To show this, we investigate four core capabilities: (i) visual quality and consistency in video predictions (Section \ref{sec:exp-vis}), (ii) the model's ability for inference-time policy improvement (Section \ref{sec:exp-improve}), (iii) the model's ability to serve as an offline simulator for policy evaluation (Section \ref{sec:exp-offline}), and (iv) the model's ability to generate multimodal conditional output distributions (Section \ref{sec:exp-multimodal}). Finally, we ablate the core components of DriftWorld in Section \ref{sec:exp-ablate}. 

\subsection{Experimental Setup}
\vspace{-2pt}

\textbf{Datasets.} We evaluate DriftWorld on five robotic manipulation datasets: the real-world Bridge-V2 \cite{walke2023bridgedata}, RT-1 \cite{rt12022arxiv}, and Language Table \cite{lynch2023interactive} datasets, as well as the simulated Push-T \cite{chi2025diffusion} and Robomimic \cite{mandlekar2021matters} datasets. We resize camera views to $256 \times 256$ for Bridge-V2 and RT-1, $192 \times 256$ for Language Table, and $96 \times 96$ for the simulated tasks.

\textbf{DriftWorld Setup.} We adapt DriftWorld's setup to the complexity of each environment. Given the current frame, 3 history frames, and $T$ future actions, DriftWorld predicts $T$ future frames. We set the prediction horizon $T$ to 4, 2, 1, 1, and 1 for the Push-T, Robomimic, Bridge-V2, RT-1, and Language Table datasets, respectively. For Push-T and Robomimic, the model generates outputs and computes the drifting loss directly in pixel space. To handle the higher resolution and complexity of the real-world datasets, the model operates in the latent space of a Stable Diffusion 3 VAE \cite{esser2024scaling,rombach2022high}, with the drifting loss computed in DINOv2/v3 \cite{oquab2023dinov2,simeoni2025dinov3} feature space.

\textbf{Baselines and Metrics.} We compare DriftWorld against the action-conditioned world models IRASim \cite{zhu2025irasim}, WorldGym \cite{quevedo2025worldgym}, Ctrl-World \cite{guo2026ctrl}, GPC \cite{qi2026inference},  and Interactive World Simulator (IWS) \cite{wang2026interactiveworldsim}. We also evaluate against three action-conditioned diffusion models formed by adapting VDM \cite{ho2022video}, LVDM \cite{he2022latent}, and AVDC \cite{ko2024learning}. These baselines include both standard and accelerated diffusion models, with WorldGym and GPC using DDIM \cite{song2021denoising} and IWS using consistency models \cite{song2023consistency} to accelerate sampling. To measure visual quality, we calculate the MSE, SSIM \cite{wang2004image}, PSNR \cite{hore2010image}, LPIPS \cite{zhang2018unreasonable}, FID \cite{heusel2017gans}, and FVD \cite{unterthiner2018towards} metrics on the validation sets. The metrics are for 8-frame autoregressive generation on Bridge-V2 and RT-1 and are for full-video autoregressive generation on Push-T, Robomimic, and Language Table.

\vspace{-5pt}
\subsection{Visual Evaluation of World Modeling}\label{sec:exp-vis}
\vspace{-3pt}

Across all five environments, DriftWorld matches or outperforms action-conditioned world model baselines in terms of visual generation quality, while operating at a fraction of the inference time.

\textbf{Push-T.} Table \ref{tab:pusht-64} displays the results for both 64-frame and full-episode ($\approx$ 250 frames) autoregressive rollouts. DriftWorld outperforms baselines while generating at $2.8\times$ to $478.8\times$ faster. Further, our model outperforms the MSE baseline model, which shares the same U-Net backbone but is trained with a standard MSE loss instead of drifting loss. Since the MSE baseline also generates frames in a single forward pass, it provides a direct comparison against DriftWorld. This comparison confirms the effectiveness of drifting loss for 1-step generation.

\textbf{Robomimic.} Table \ref{tab:robomimic-all} presents the visual quality metrics on the Robomimic tasks. In both single-view and two-view settings, DriftWorld achieves competitive performance compared to Ctrl-World and GPC world models, and it is significantly faster.

\begin{figure*}[t]
    \centering
    \includegraphics[width=\linewidth]{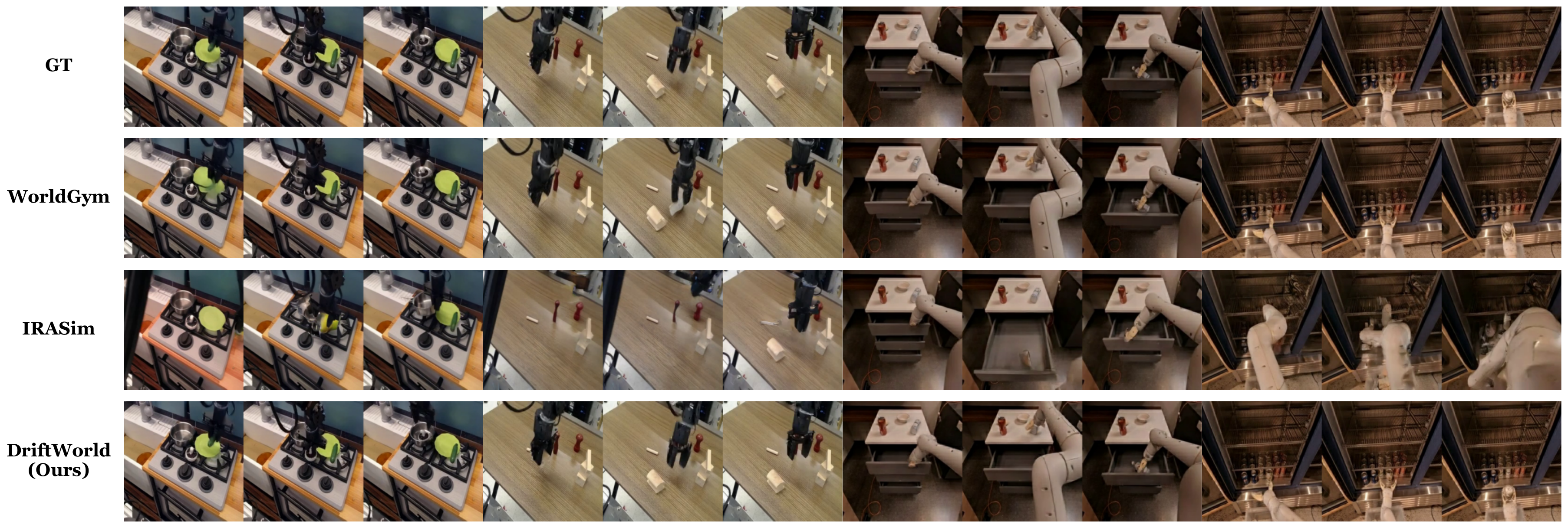}
    \vspace{-15pt}
    \caption{\textbf{Comparison of world-model rollouts on Bridge-V2 and RT-1.} DriftWorld generates videos that accurately follow action conditioning and real-world dynamics. By using a single forward pass, it achieves this high-quality generation at a fraction of the inference time required by baseline models. Due to space limitations, the initial conditioning frame is omitted, and only generated frames are shown. Full videos are on our anonymous website.
    }
    \label{fig:bridge-rt1}
    \vspace{-12pt}
\end{figure*}

\textbf{Bridge-V2, RT-1, and Language Table.} As Table \ref{tab:bridge} shows, DriftWorld outperforms baselines on the majority of the visual quality metrics, while operating $11.8\times$ to $162.5\times$ faster than IRASim, LVDM, and VDM. Compared with the similarly fast Interactive World Simulator baseline, DriftWorld achieves higher visual quality, with much lower FID and FVD. Generated videos are visualized in Figure \ref{fig:bridge-rt1}. DriftWorld produces visually sharp rollouts that closely follow the action conditioning and the dynamics of the real-world environment.

\begin{table}[!t]
    \centering
    \vspace{2pt}
    \tiny
    \setlength{\tabcolsep}{3.2pt}
    \renewcommand{\arraystretch}{1.03}
    \resizebox{\linewidth}{!}{%
    \begin{tabular}{lccccc@{\hspace{6pt}}ccccc}
        \toprule
        & \multicolumn{5}{c}{64-frame rollouts} & \multicolumn{5}{c}{Full-episode rollouts} \\
        \cmidrule(lr){2-6}
        \cmidrule(lr){7-11}
        World Model & MSE $\downarrow$ & SSIM $\uparrow$ & PSNR $\uparrow$ & LPIPS $\downarrow$ & Time (s) & MSE $\downarrow$ & SSIM $\uparrow$ & PSNR $\uparrow$ & LPIPS $\downarrow$ & Time (s) \\
        \midrule
        GPC \cite{qi2026inference} & 0.0033 & 0.9717 & 33.1086 & 0.0239 & 0.0104 & \textbf{0.0033} & 0.9713 & \textbf{32.3692} & 0.0278 & 0.0309 \\
        AVDC \cite{ko2024learning} & 0.0045 & 0.9862 & 30.4990 & 0.0100 & 0.1558 & 0.0063 & 0.9825 & 29.8518 & 0.0143 & 0.4897 \\
        Ctrl-World \cite{guo2026ctrl} & 0.0540 & 0.8914 & 18.0705 & 0.1844 & 1.7714 & 0.0576 & 0.8854 & 18.0638 & 0.2513 & 2.0154 \\
        MSE Baseline & 0.0105 & 0.9704 & 28.0578 & 0.0225 & \textbf{0.0037} & 0.0150 & 0.9632 & 26.5779 & 0.0389 & \textbf{0.0045} \\
        DriftWorld (Ours) & \textbf{0.0020} & \textbf{0.9941} & \textbf{34.7751} & \textbf{0.0035} & \textbf{0.0037} & 0.0061 & \textbf{0.9869} & 32.6608 & \textbf{0.0124} & \textbf{0.0045} \\
        \bottomrule
    \end{tabular}%
    }
    \caption{\textbf{Visual quality of generated videos for Push-T}, averaged over 1000 seeds for the starting environment. We also report the average generation time (seconds per frame) for each world model on a single H100 GPU.}
    \label{tab:pusht-64}
    \vspace{-4pt}
\end{table}

\begin{table}[!t]
    \centering
    \vspace{2pt}
    \tiny
    \setlength{\tabcolsep}{3.2pt}
    \renewcommand{\arraystretch}{1.03}
    \resizebox{\linewidth}{!}{%
    \begin{tabular}{lccccc@{\hspace{6pt}}ccccc}
        \toprule
        & \multicolumn{5}{c}{1-view} & \multicolumn{5}{c}{2-view} \\
        \cmidrule(lr){2-6}
        \cmidrule(lr){7-11}
        Model & MSE $\downarrow$ & SSIM $\uparrow$ & PSNR $\uparrow$ & LPIPS $\downarrow$ & Time (s) & MSE $\downarrow$ & SSIM $\uparrow$ & PSNR $\uparrow$ & LPIPS $\downarrow$ & Time (s) \\
        \specialrule{\lightrulewidth}{\aboverulesep}{0pt}
        \rowcolor{gray!20} & \multicolumn{5}{l}{\emph{Lift task}} & \multicolumn{5}{l}{\emph{Lift task (2-view): wrist view}} \\
        GPC \cite{qi2026inference} & 0.0152 & \textbf{0.9306} & 27.7798 & \textbf{0.0241} & 0.0331 & 0.0291 & 0.9376 & 25.1058 & 0.0770 & 0.0374 \\
        Ctrl-World \cite{guo2026ctrl} & 0.0334 & 0.8698 & 22.4874 & 0.0504 & 3.1602 & 0.0650 & 0.9020 & 19.8061 & 0.1724 & 3.1619 \\
        DriftWorld (Ours) & \textbf{0.0071} & 0.8907 & \textbf{30.4492} & 0.0401 & \textbf{0.0100} & \textbf{0.0122} & \textbf{0.9560} & \textbf{29.3393} & \textbf{0.0358} & \textbf{0.0100} \\
        \specialrule{\lightrulewidth}{\aboverulesep}{0pt}
        \rowcolor{gray!20} & \multicolumn{5}{l}{\emph{Can task}} & \multicolumn{5}{l}{\emph{Lift task (2-view): agent view}} \\
        GPC \cite{qi2026inference} & 0.0272 & \textbf{0.8917} & 22.9822 & 0.0711 & 0.0331 & 0.0230 & 0.9079 & 25.4217 & 0.0352 & 0.0374 \\
        Ctrl-World \cite{guo2026ctrl} & 0.0246 & 0.8801 & 22.6784 & 0.0817 & 3.1602 & 0.0458 & 0.8181 & 19.7703 & 0.0905 & 3.1619 \\
        DriftWorld (Ours) & \textbf{0.0096} & 0.8858 & \textbf{28.8460} & \textbf{0.0614} & \textbf{0.0100} & \textbf{0.0102} & \textbf{0.9282} & \textbf{28.8281} & \textbf{0.0258} & \textbf{0.0100} \\
        \bottomrule
    \end{tabular}%
    }
    \caption{\textbf{Visual quality of generated videos on Robomimic tasks.} The left half is for single-view generation on the Lift and Can tasks, and the right half is for two-view generation (agent and wrist views) on the Lift task.}
    \label{tab:robomimic-all}
    \vspace{-3pt}
\end{table}

\begin{table}[!t]
    \centering
    \scriptsize
    \begin{tabular}{lccc}
        \toprule
        World Model & Policy 1 IoU & Policy 2 IoU & Time (s) \\
        \midrule
        N/A (Base Policy) & 0.635 & 0.612 & - \\
        \midrule
        GPC World Model \cite{qi2026inference} & 0.698 & 0.614 & 2.241 \\
        AVDC \cite{ko2024learning} & 0.726 & 0.682 & 106.1 \\
        MSE Baseline & 0.639 & 0.609 & \textbf{0.912} \\
        DriftWorld (Ours) & \textbf{0.772} & \textbf{0.755} & \textbf{0.912} \\
        \bottomrule
    \end{tabular}
    
    \caption{\textbf{GPC-RANK Performance.} For two policies, this table shows its IoU score after applying GPC-RANK with each world model, compared to its IoU score without GPC-RANK. We record the time taken for each world model to roll out all $K$ action proposals.}
    \label{tab:gpc-rank}
    \vspace{-4pt}
\end{table}

\begin{table*}[t]
    \centering
    \scriptsize 
    \begin{tabular}{clcccccc}
        \toprule
        Dataset & World Model & SSIM $\uparrow$ & PSNR $\uparrow$ & LPIPS $\downarrow$ & FID $\downarrow$ & FVD $\downarrow$ & Time (s) \\
        \midrule
        \multirow{5}{*}{Bridge-V2} 
        & IRASim \cite{zhu2025irasim} & 0.738 & 18.836 & 0.158 & 10.46 & 91.55 & 1.1031 \\
        & IWS \cite{wang2026interactiveworldsim} & 0.786 & \textbf{22.847} & 0.169 & 45.30 &	329.75 & 0.0255 \\
        & LVDM \cite{he2022latent} & 0.741 & 19.191 & 0.154 & 9.86 & 103.51 & 0.2850 \\
        & VDM \cite{ho2022video} & 0.725 & 18.417 & 0.167 & 39.88 & 159.16 & 3.2078 \\
        & DriftWorld (Ours) & \textbf{0.835} & 22.691 & \textbf{0.091} & \textbf{4.86} & \textbf{46.05} & \textbf{0.0242} \\
        
        \midrule
        \multirow{5}{*}{RT-1}
        & IRASim \cite{zhu2025irasim} & 0.757 & 21.053 & 0.151 & \textbf{4.60} & 93.02 & 1.1043 \\
        & IWS \cite{wang2026interactiveworldsim} & 0.807 & \textbf{24.286} & 0.167 & 46.81 & 448.21 & 0.0229 \\
        & LVDM \cite{he2022latent} & 0.750 & 20.996 & 0.153 & 5.40 & 94.58 & 0.2844 \\
        & VDM \cite{ho2022video} & 0.540 & 13.822 &	0.246 & 31.68 & 406.13 & 3.1527 \\
        & DriftWorld (Ours) & \textbf{0.841} & 24.203 & \textbf{0.097} & 8.99 & \textbf{62.84} & \textbf{0.0194} \\

        \midrule
        \multirow{5}{*}{Language Table}
        & IRASim \cite{zhu2025irasim} & 0.877 & 26.894 & 0.055 & 9.56 & 61.19 & 1.1381 \\
        & IWS \cite{wang2026interactiveworldsim} & 0.903 & 25.956 &	0.090 & 20.81 & 160.51 & 0.0182 \\
        & LVDM \cite{he2022latent} & 0.874 & 26.274 & 0.064 &  10.03 & 38.09 & 0.2704 \\
        & VDM \cite{ho2022video} & 0.796 & 21.455 & 0.140 & 26.76 & 43.60 & 0.8660 \\
        & DriftWorld (Ours) & \textbf{0.941} & \textbf{29.596} & \textbf{0.045} & \textbf{8.89} & \textbf{22.96} & \textbf{0.0166} \\
        \bottomrule
    \end{tabular}

    \vspace{0.2cm}
    
    \caption{\textnormal{\textbf{Visual quality of generated videos on the real-world Bridge-V2, RT-1, and Language Table datasets.} We also report the average generation time (seconds per frame) for each world model.}}
    \label{tab:bridge}
    \vspace{-5pt}
\end{table*}

\vspace{-2pt}
\subsection{Inference-Time Policy Improvement in DriftWorld}\label{sec:exp-improve}
\vspace{-3pt}

DriftWorld enables faster and more effective inference-time policy improvement compared to existing diffusion world models. We evaluate this capability by taking a pretrained policy, rolling out its action proposals in DriftWorld, and selecting the actions that result in the best future states.

In particular, we employ the GPC-RANK approach \cite{qi2026inference}. At each step, the base policy samples $K$ candidate action proposals. Each proposal is rolled out in imagination through the world model. Then, a reward function (a ResNet18+MLP model from \cite{qi2026inference}) scores the predicted future observations, and the action chunk with the highest reward is selected to execute in the environment.

We apply this method to two diffusion policies trained for different numbers of epochs \cite{chi2025diffusion}. As shown in Table \ref{tab:gpc-rank}, applying GPC-RANK with $K=50$ using DriftWorld significantly boosts the final Intersection over Union (IoU) score of both policies compared to the base policies. Moreover, DriftWorld achieves better performance than the GPC world model, AVDC, and an MSE baseline, while rolling out all candidate proposals in a fraction of the inference time.

\vspace{-4pt}
\subsection{DriftWorld as an Offline Simulator for Policy Evaluation}\label{sec:exp-offline}
\vspace{-2pt}

DriftWorld serves as an accurate offline simulator for policy evaluation, successfully predicting both the absolute performance and relative ranking of policies. We validate this capability across both the Robomimic and Push-T tasks.

\begin{figure}[h!]
    \centering
    \includegraphics[width=\linewidth]{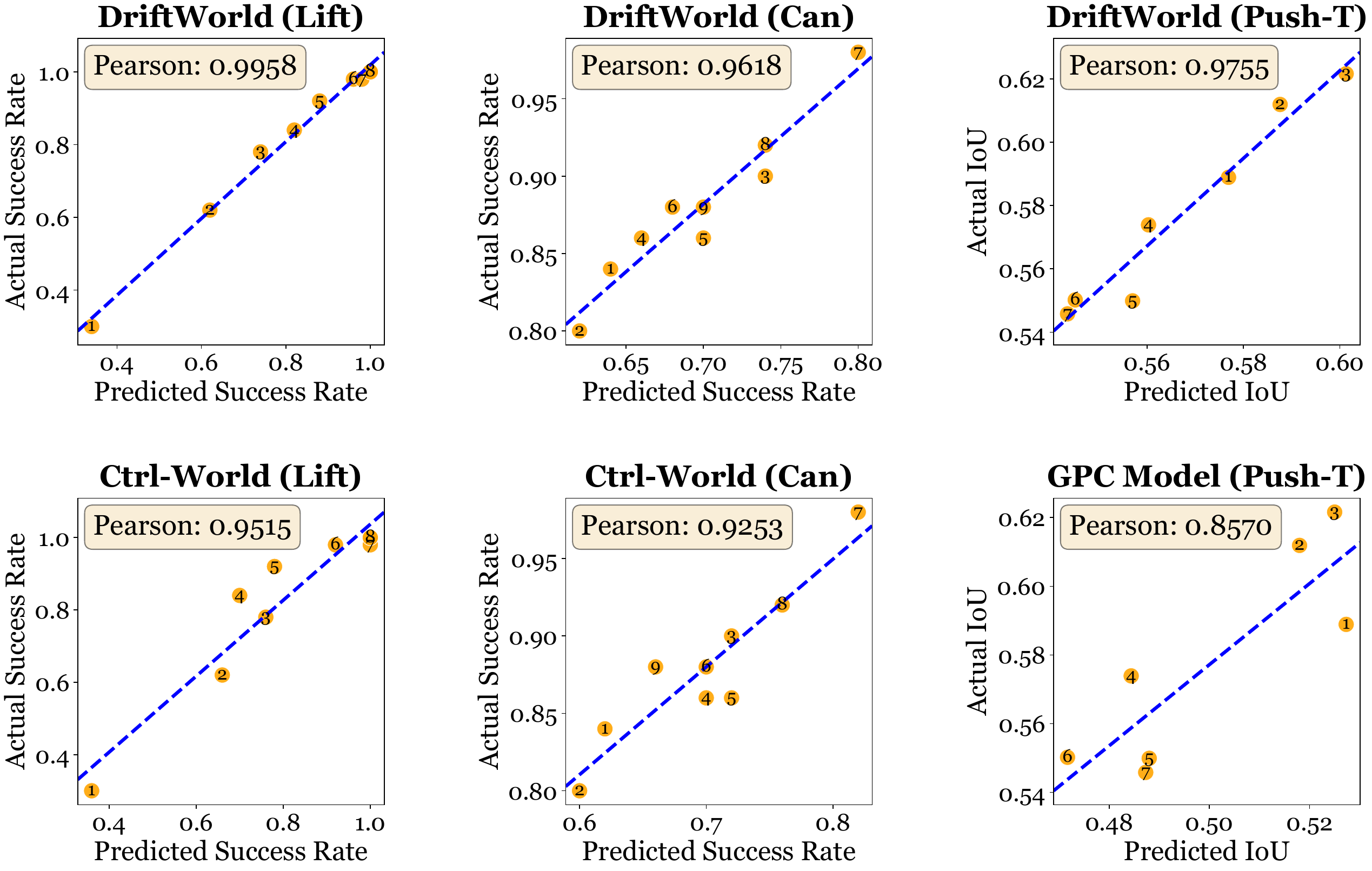}
    \vspace{-17pt}
    \caption{\textbf{Policy Evaluation.} On the Lift, Can, and Push-T tasks, DriftWorld achieves a higher correlation coefficient between predicted and ground-truth values of the success rates or IoU scores than the baseline.}
    \label{fig:policy-eval-both}
    \vspace{-2pt}
\end{figure}

\textbf{Push-T.} We evaluate seven diffusion policies \cite{chi2025diffusion}, comparing their average IoU scores across 100 seeds in DriftWorld versus in the ground-truth simulator. As Figure \ref{fig:policy-eval-both} shows, DriftWorld correctly ranks all policies except for one pair and achieves a high Pearson correlation coefficient of 0.9755. In contrast, the baseline GPC model \cite{qi2026inference} has several ranking errors and a lower correlation of 0.8570.

\textbf{Robomimic.} We further validate DriftWorld on the Robomimic tasks. We roll out nine diffusion policies \cite{chi2025diffusion} in the world model and record average success rates across 50 environment initialization seeds. The simulated success rates in DriftWorld closely mirror those in the ground-truth MuJoCo simulator with correlation coefficients of 0.9958 and 0.9618, outperforming the baseline Ctrl-World model \cite{guo2026ctrl}.

\vspace{-5pt}
\begin{table*}[t]
    \centering
    \tiny
    \setlength{\tabcolsep}{2.5pt}
    \begin{tabular}{@{}cccccc@{\hspace{3em}}cccccc@{\hspace{3em}}cccccc@{}}
        \toprule
        \multicolumn{6}{c}{Ablation on Feature Extractor} & \multicolumn{6}{c}{Ablation on Action Accentuation $\alpha$} & \multicolumn{6}{c}{Ablation on Improvements for Autoregressive Generation} \\
        \cmidrule(r{3em}){1-6}\cmidrule(r{3em}){7-12}\cmidrule{13-18}
        Extractor & SSIM $\uparrow$ & PSNR $\uparrow$ & LPIPS $\downarrow$ & FID $\downarrow$ & FVD $\downarrow$ & $\alpha$ & SSIM $\uparrow$ & PSNR $\uparrow$ & LPIPS $\downarrow$ & FID $\downarrow$ & FVD $\downarrow$ & Improvements & SSIM $\uparrow$ & PSNR $\uparrow$ & LPIPS $\downarrow$ & FID $\downarrow$ & FVD $\downarrow$ \\
        \midrule
        None & 0.856 & \textbf{25.946} & 0.107 & 34.24 & 168.34 & None & 0.816 & 22.747 & 0.124 & 22.03 & 191.05 & None & 0.824 & 22.320 & 0.120 & \textbf{8.73} & 174.67 \\
        DINOv2 & \textbf{0.892} & 25.059 & 0.047 & \textbf{0.40} & 13.58 & 2.5 & \textbf{0.841} & \textbf{24.203} & \textbf{0.097} & \textbf{8.99} & \textbf{62.84} & Motion Weight & 0.830 &	23.416 & 0.109 & 11.63 & 67.88 \\
        DINOv3 & \textbf{0.892} & 25.000 & \textbf{0.046} & 0.62 & \textbf{6.20} & 4 & 0.836 & 23.852 & 0.101 & 9.45 & 63.72 & Motion Weight + SF & \textbf{0.841} & \textbf{24.203} & \textbf{0.097} & 8.99 & \textbf{62.84} \\
        \bottomrule
    \end{tabular}
    
    \caption{\textbf{Ablations on core components in DriftWorld.} Using either DINOv2 or DINOv3 as a feature extractor significantly improves performance on Bridge-V2 (left). Using action accentuation is also beneficial (middle). Incorporating motion weighting and self-forcing improve DriftWorld's performance on autoregressive generation on RT-1 (right).}
    \label{tab:ablation-full}
    \vspace{-10pt}
\end{table*}

\subsection{Multimodal Generation with DriftWorld}
\label{sec:exp-multimodal}
\vspace{-5pt}

The following experiments show that DriftWorld can learn multimodal conditional output distributions: for a fixed conditioning, the model produces multiple distinct, valid futures when given different noise inputs. To evaluate model performance on multimodal generation, we use the language-conditioned setting in which the model generates future frames given an initial observation $o_0$ and a language instruction $\ell$, e.g. ``slide the blue cube above the blue triangle." This setting is more suitable for evaluating multimodal generation performance than the main action-conditioned setting. This setting provides a more natural test of multimodal generation than the main action-conditioned setting: a single language instruction can be accomplished through different gripper trajectories, while under action conditioning the future path is largely determined, up to small real-world uncertainty.

\begin{figure}[h!]
    \vspace{-6pt}
    \centering
    \includegraphics[width=\linewidth]{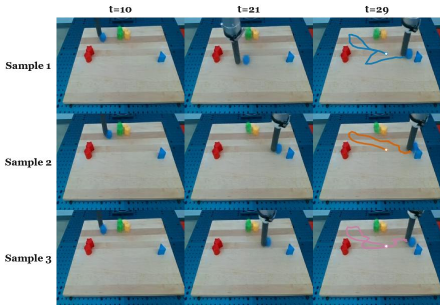}
    \vspace{-15pt}
    \caption{\textbf{Multimodal language-conditioned generation.} Given the same instruction ``slide the blue cube above the blue triangle," DriftWorld generates videos showing multiple valid trajectories (highlighted in the last frame).}
    \label{fig:multimodal}
    \vspace{-10pt}
\end{figure}

Figure \ref{fig:multimodal} shows that given the same instruction and initial frame, DriftWorld generates videos showing multiple valid trajectories. Table \ref{tab:multimodal} further quantifies this behavior using the diversity metrics of Vendi-DINO \cite{friedman2023vendi} and gripper endpoint spread. While DriftWorld has slightly lower diversity scores than the diffusion baselines, it achieves competitive best-of-16 visual quality.

These results show that training with our single-positive drifting objective (Section \ref{sec:drift-1pos}) does not cause the model to collapse to a deterministic point estimate, but instead allows it to recover multiple plausible modes of the data distribution.

\begin{table}[h!]
    \centering
    \tiny
    \setlength{\tabcolsep}{4pt}
    \begin{tabular}{lcccccccc}
        \toprule
        Model & Vendi $\uparrow$ & Endpoint Spread $\uparrow$ & SSIM $\uparrow$ & PSNR $\uparrow$ & LPIPS $\downarrow$ & FID $\downarrow$ & FVD $\downarrow$ \\
        \midrule
        IRASim & 2.42 & 23.86 & 0.853 & \textbf{24.190} & 0.099 & \textbf{18.27} & 120.34 \\
        LVDM & \textbf{2.52} & 22.16 & 0.842 & 23.107 & 0.104 & 18.43 & 135.58 \\
        VDM & 2.36 & \textbf{26.59} & 0.773 & 21.666 & 0.164 & 45.81 & 262.34 \\
        DriftWorld & 2.04 & 14.01 & \textbf{0.861} & 23.014 & \textbf{0.094} & 21.30 & \textbf{115.16} \\
        \bottomrule
    \end{tabular}
    
    \caption{\textbf{Diversity and best-of-16 visual quality metrics.} DriftWorld achieves multimodal generation while achieving competitive visual quality relative to diffusion baselines.}
    \label{tab:multimodal}
    \vspace{-5pt}
\end{table}

\vspace{-3pt}
\subsection{Ablations}\label{sec:exp-ablate}
\vspace{-2pt}

Finally, we ablate several core components of DriftWorld.

\textbf{Ablation on feature space.} Table \ref{tab:ablation-full} ablates the feature space where drifting loss is computed. The metrics are for frames generated by one forward pass of DriftWorld on the validation set. Without a feature extractor, performance drops significantly and the gripper becomes blurry. In contrast, using DINOv2/v3 leads to sharp generated videos.

\textbf{Ablation on accentuated action conditioning.} As Table \ref{tab:ablation-full} shows, using accentuated action conditioning greatly improves model performance. The metrics are on RT-1. Performance varies slightly among different accentuation scales $\alpha$, with $\alpha=2.5$ performing better in this case.

\textbf{Ablation on motion weighting and self-forcing.} Table \ref{tab:ablation-full} shows both components are crucial for DriftWorld's autoregressive generation on real-world datasets. Motion weighting allows DriftWorld to precisely capture gripper movements, significantly reducing FVD, because the drifting loss is weighted more heavily on moving regions than the static background. Adding a second self-forcing training stage \cite{huang2025self}, where the model is conditioned on its own predictions rather than ground-truth frames, further improves performance.

\section{Discussion}

In this paper, we introduced DriftWorld, a fast, single-step, action-conditioned world model. To adapt drifting generative models to conditional video generation, we proposed several key adaptations, including a single-positive drifting objective, feature space integration, and accentuated action following. DriftWorld achieves generation speeds over 40 fps, which is a 12+ times speedup over state-of-the-art diffusion baselines, while producing accurate and visually consistent rollouts.
\vspace{-2pt}

\bibliographystyle{IEEEtran}
\bibliography{main}

\end{document}